# One-Shot Item Search with Multimodal Data

Jonghwa Yim
Samsung Electronics
jonghwa.yim@samsung.com

Junghun James Kim
Samsung Electronics
jameshun@samsung.com

Daekyu Shin
Samsung Electronics
daekyu.shin@samsung.com

## Abstract

*In the task of near similar image search, features from Deep Neural Network is often used to compare images and measure similarity. In the past, we only focused visual search in image dataset without text data. However, since deep neural network emerged, the performance of visual search becomes high enough to apply it in many industries from 3D data to multimodal data. Compared to the needs of multimodal search, there has not been sufficient researches.*

*In this paper, we present a method of near similar search with image and text multimodal dataset. Earlier time, similar image search, especially when searching shopping items, treated image and text separately to search similar items and reorder the results. This regards two tasks of image search and text matching as two different tasks. Our method, however, explore the vast data to compute k-nearest neighbors using both image and text.*

*In our experiment of similar item search, our system using multimodal data shows better performance than single data while it only increases minute computing time. For the experiment, we collected more than 15 million of accessory and six million of digital product items from online shopping websites, in which the product item comprises item images, titles, categories, and descriptions. Then we compare the performance of multimodal searching to single space searching in these datasets.*

## 1. Introduction

Since deep neural network emerged, in many IT industries, there have been many researches regarding near similar image search. For example, Google proved similar image search on a web base on large scale dataset. Pinterest, eBay, and SK Planet [1]–[3] also provided similar search in fashion item dataset. Yet, the precision of similar search is not enough and has a lot of constraints in the environment.

Feature fusion in the task of similar image search [4]–[7] is recently being researched and greatly improved the technology. Earlier approaches attempted to extract multiple features in a given query image such as SIFT, LBP, HOG and color descriptor. The problem is that these are image-to-image searching methods. Also, we do not know which features are necessary for a given task in real industry. On top of that, these are extension of image feature extraction and not suitable for multimodal search.

Not like similar image search in small space, finding similar images in enormously large dataset in the blink of an eye is a difficult problem and often leads failure of finding the best answer in the database. Since brute force algorithm is not appropriate for a huge dataset, there are some researches [8], [9] how to build a system to make searching item in the dataset efficiently. These methods are based on distance measurements to build navigable small worlds. To fully accommodate fast-searching method, we have to prevent increasing computing time of distance metric in custom dataset of industries. This becomes especially harder on multimodal dataset. For years, many systems [10]–[18] were developed for efficient search of items in large multimodal dataset. Most of them use image and text feature independently by searching twice, and merge two results and rerank in some logical way. These methods lead inefficiency by increasing searching time and/or not finding the best nearest neighbor. Also it heavily depends on how to merge and reorder two results.

In this paper, we vectorize text and image data first for efficient search of multimodal data. And then we concatenate two features to configure augmented feature vector and build navigable world. For query images, we first encode images by feeding into the neural network, extract words from images using image classification, vectorize label words, and then concatenate image and word vectors. Since the length of augmented vector of query image is matched to that of target dataset, we can search nearest neighbors in preconfigured large dataset.

## 2. Related Work

Recently there are a few efforts on visual search, especially in the industry of web shopping. Previous research by Pinterest experimented visual search architecture on fashion items in Pinterest by detecting target object to extract target feature precisely. It uses text data to predict the image category to determine whether to apply object detection module specific to the predicted category.

## 2.1. Visual Search

Pinterest, Google, Bing, Alibaba, Amazon Flow are examples of visual search widely used. All of them uses visual features with their own architecture, and for some of them, text info is also used to get a satisfactory result of the visual search, as shown in [1], [15]–[17]. Also, various works such as [19], [20] were done to improve the ranking system using visual features. Compared with existing commercial visual search engines, our system focuses more on getting fine results by using not only the image features but also the text info from the product image.

## 2.2. Multimodal Search.

There is previous work [18] aiming multimodal search for shopping, especially fashion and furniture items in large dataset. This method proposes DeepStyle network for feature fusion which uses visual and textual vector to generate multimodal vector representation, which is the major difference from our method. Also this method uses context space to train the multimodal vector to learn context of image, category-like output, which weakens near similarity search. The Late-fusion Blending and Early-fusion Blending in there is also largely different from our method in that this blending method is a kind of sequential search, instead of one-shot search in our method. Instead of using extra network or context learning, our method is simple but strong multimodal search. Moreover, compared to previous method, our method can adjust the weight of textual vector to visual vector, enabling us to customize the output. Additionally, previous researches [18], [21] require user input text dictating the desired attribute. However, our method does not necessarily require user input text.

## 2.3. Feature Representation.

Previous works on visual search [1], [15]–[17], [22] have used visual features for classification engines, such as category classifier. They also have used the visual features to get candidate results by using methods, distance measure like Euclidean or Manhattan distance metric and clustering techniques like K-means. We also use visual features to get the classification like the others have done. However, we have approached the feature matching, getting the candidate results, in a different way. We have trained visual feature model using near similar metric learning method and vectorized textual features from the class label to use it with visual feature vector, while others use only visual features. In other words, our feature representation is not like ones from previous works.

## 3. Features for Nearest Neighbor Search

## 3.1. Visual Features

In the experiment, we treat large amount of images. With acceptable and reasonable computing time, we need to extract good image representations. We have applied small modification in Inception-v1 [22] network, resizing the final concatenated feature to 1024-dimension after removing the top fully connected layer as well as other middle layers down. Concatenated 1024-dim feature vector from image $v_I$ is a base feature representation for similar item search in this paper. Like previous researches [1]–[3], for both target dataset and query image, we used the same feature representation for consistency. To train the neural network model to learn similarity feature embedding where similar images are mapped close and dissimilar images away from each other, we use previous research [23] to train the network, in order to detect similar images versus negative samples. With this network, we can pull out similar items from shopping item database, even when the same items not exist.

## 3.2. Classifier

For better visual search, we use visual classifier to detect specific category of target object. The labels can be specific category of the object, like "ring", "necklace", etc. Or we can also label and train the attribute of the items, like "round neck", "v-neck", "collar", etc., and merge them to get better results from visual search. In many previous methods [2], [3], [12], [13], we often use category-like labels to match to the database items or re-rank the result of visual search. These methods strongly depend on customization. Also, these post-processing methods are slow and inaccurate.

Our method, however, uses the classification labels at the time of visual search so that we can use a cosine similarity metric, making nearest neighbor search one-shot, fast, accurate and efficient. In the following section we describe how we use the labels to generate multimodal feature vector. Our method uses 59 category labels for accessory items and 58 category labels for digital items from shopping mall. Therefore, for every input image I, we put it into the classifier $f(I)$, and get label $l_{cls} \in \{l_1, \ldots, l_k\}$, then we vectorize it with one of two different methods, one hot encoding and textual vectorization, for visual search.

## 3.3. Visual Features from Classifier

Previous research [24] proves that features extracted from a deep neural network trained in a supervised manner can be clustered well and perform as a representative feature for generic tasks. Since we are using labels from classifier to generate word vector, we experiment with features from classifier network to form database and query image feature, which is a form of vector in 2048-dimension collected from the last pooling layer in classifier. We also use this feature to search the nearest neighbor items.

### 3.4. Textual Features

Recently, in many industries, especially shopping or social network, they require similar item search based on image and text data. Thus, instead of focusing just on the visual features, we considered to use text data for word features. Text data is related to its images in the way that it expresses the title, description, category of the image, written or defined by image provider, such as merchants of shopping mall or user of social network. The combination can be title text $t_1$ plus category text $t_2$ or title text $t_1$ plus description text $t_3$. Sometimes, there are misallocated categories, typos, and wrongly written texts in product text data from shopping mall. Although there are some errors and typos in the text data, we did not manually correct it, and our experiment shows reasonably good results in the following evaluation.

To compute text vector representation, we used the title and category texts for shopping mall dataset. Then we removed stop words, numbers, and symbols other than the English characters. Refined texts from $t_1$ and $t_2$ are then concatenated in row to make a single sentence. Previous researches [25]–[27] introduced word-to-vector methods. These methods project the words onto the vector space in the way to preserve the similarity of words in the vector space. So the cosine similarity of the vectors represents the words similarity. Especially, word to vector method of Facebook research [27] utilize n subwords so that it can also calculate proper word vectors from misspelled words. Since our datasets inherently include some misspelled words and words not in English dictionary, we used this method of word to vector with subword information.

To vectorize text data from an item, we applied a simple rule to title and category path. Firstly, to remove meaningless words such as 'the', 'a', etc., we removed stop-words, symbols, and also numbers. Then we get the sentence vector $\mathbf{v_W}$ by vectorizing each words and normalize. Let's assume that refined sentence consists of n words. For every $\mathbf{v_k}$ in the sentence,

$$\mathbf{v_W} = \frac{1}{n} \sum_{k=1}^{n} \frac{\mathbf{v_k}}{\|\mathbf{v_k}\|} \quad (1)$$

Using cosine similarity metric, we compute the similarity of words by the relational direction, not with the length of the vectors. Therefore, normalization is a good starting point to get the vector from sentence. Otherwise, we could normalize after sum up all vectors from words to get sentence vector $\mathbf{v_W}$:

$$\mathbf{v_W} = \frac{\frac{1}{n}\sum_{k=1}^{n} \mathbf{v_k}}{\left\|\frac{1}{n}\sum_{k=1}^{n} \mathbf{v_k}\right\|} \quad (2)$$

for *n* words in the sentence. As generally known, parameter normalization usually gives us better result and it also makes sense in our experiment because we are calculating the angle of sentence vector, and considering that the length is not meaningful. We also empirically found out that the normalization before summing up gives us slightly better result in the further experiment.

For the final step, we trained word vectorizing model using the text data from shopping mall. The text data for training contains concatenated texts of category, title, and description. After removing stop words and counting words shown more than five times, we left around 977k words to train the model. Since the number of words in our training dataset is lesser than that from original paper, three million, we set the textual vector to be 110-dimensional vector, slightly lower dimension compared to the original paper. We also empirically verified that no larger vector dimension is needed.

### 3.5. Similarity Metric

To find nearest neighbor, we must define how to measure the distance between each data. Within the limited time, searching in the vast dataset requires efficient logic other than the brute-force algorithm. To build an efficient searching system, we adopted previous research [9], which builds small navigable worlds for efficient searching. Here and there, we use cosine similarity to measure the similarity distance d of two vectors.

$$d = \frac{(\mathbf{v_1} \cdot \mathbf{v_2})}{|\mathbf{v_1}| \cdot |\mathbf{v_2}|} \quad (3)$$

This metric is useful for us to calculate vectorized neighbors' distances. Thus, we use cosine similarity to explore the nearest neighbors in the dataset.

## 4. One-shot Multimodal Search

### 4.1. One-hot Encoded Augmentation

In this paper, we target multimodal item search using class labels. To search items in the database consisted of images and texts, at first, we manually classified database items to category label which is the output of input image classifier. So we forward-pass the input image into the neural network classifier to get the class label y. Then we wanted to use this label to find near similar items of corresponding class. To do so, we experimented simple but effective vectorizing method, one-hot encoded vector $\boldsymbol{\delta}$

using class label:

$$\delta_i = \begin{cases} 1 \text{ if } i = \underset{i}{\operatorname{argmax}} \frac{e^{y_i}}{\sum_{k=1}^{K} e^{y_k}} \\ 0 \text{ otherwise} \end{cases} \quad (4)$$

where $\delta_i \in \boldsymbol{\delta} = [\delta_1, ..., \delta_K]$. Additionally we allocate $\delta_0$ for unknown class. This one-hot encoded vector enables near similar search to find items of same category label. We fuse $\boldsymbol{\delta}$ with image vector $\mathbf{v_I}$ to form multimodal vector. Cosine similarity is used for similar item search. Therefore, products with similar image but different item tend to be removed from the search results. In some cases, classification error yields wrong search results. To relieve the side-effect, we then apply textual vector augmentation.

### 4.2. Multimodal Feature Augmentation

From the multimodal dataset of ours, we extract two vectors $\mathbf{v_I}$, $\mathbf{v_W}$, each from image and text. To find similar item in multimodal dataset, some previous methods [12]–[14] used sequential search: image search first, text search next, and finally merging two results by their own rules. On the other hand, it is also possible to add two normalized vectors of same dimension together to get a summed vector. Be sure to notice that once we get a single multimodal vector representation, the great advantage is that we can make nearest neighbor search one-shot using cosine similarity metric to search in huge data space. Likewise, image feature vector can be added on the text feature vector to generate single feature vector for nearest-neighbor search with one-shot. However, this cannot be guaranteed if each corresponding pair of elements of two vectors is originated from the different training data and different space. Therefore, in this paper, to make one shot searching system using cosine similarity metric, we concatenate two vectors, $\mathbf{v_I}$ and $\mathbf{v_W}$, to get multimodal vector $\mathbf{v_{aug}}$:

$$\mathbf{v_{aug}} = \mathbf{v_I} \oplus \mathbf{v_W} \quad (5)$$

, so that the cosine similarity metric now becomes like

$$d = \frac{(\mathbf{v_{aug}^1} \cdot \mathbf{v_{aug}^2})}{|\mathbf{v_{aug}^1}| \cdot |\mathbf{v_{aug}^1}|} \quad (6)$$

and we measured the performance of similar item search.

Especially, we can adjust the weight of textual vector versus visual vector by normalizing and multiplying weight factor w. Then the multimodal vector now becomes:

$$\mathbf{v_{aug}} = \mathbf{v_I} \oplus (\mathbf{v_W} \times w) \quad (7)$$

and this weight factor can also be used for one-hot vectored augmentation method in previous section (section **4.2**), making experimental customization handy.

## 5. Experiments

### 5.1. Dataset

In this paper, our contribution is a fast and efficient one-shot nearest-neighbor searching system with an augmented feature vector in multimodal database. For practical purpose, we collected image and text data dataset from online shopping malls to set multimodal database. Shopping item is most common multimodal data of various images and texts. Moreover, multimodal searching system is highly demanded in shopping industry. Instead of using public dataset such as Fashoin200 [28] or DeepFashion [29], which only contain scant text data and far from real shopping mall texts, we internally gathered items from web shopping mall. Since the number of existent shopping item is too large, we aimed the target to accessory and digital item. We collected the accessory and digital shopping item around ~650k and ~380k for each. One item includes item image, title, and category path provided by merchandiser.

### 5.2. Evaluation

It is hard to objectively score the result of similar item search in the dataset of similar items. Moreover, the size of the database is too large to manually mark the most similar items. Thus, in our experiment, five human raters recorded the top-5 score by manually counting the number of similar products within the top-5 results. To compare the results and score the accuracy, we placed four images - result without feature augmentation (section **3.1**), result using the feature from the classifier model (section **3.3**), result with feature augmentation using one-hot encoding from classifier (section **4.1**), and result with feature augmentation using word vector (section **4.2**) - and marked the number of similar items found from the top-5. Then we calculated the average score to compare the results.

## 6. Results

First, we have tested our methods on the database of accessory items gathered from web shopping mall. Testset contains 120 images collected from multiple websites and customer reviews. Based on the evaluation method described in section **5.2**, we have marked the average top-5 score, in Table 1. In this table, 'Original' stands for the score from the network which is trained to learn similarity metric (section **3.1**) as a feature extractor. The score by using the classifier network is marked 'Classifier.' The score by using the multi-network, combining the network from section **3.1** with a one-hot encoding augmentation (section **4.1**), is marked 'Ours$_1$.' The score by using the method, combining the image feature from section **3.1** with word vector, is marked 'Ours$_2$.' The length of our concatenated vector is about 10% larger than the original vector. Despite of its minor disadvantage in the size, the score is highly increasing. Ours$_2$ especially shows great performance gain while Ours$_1$ is fairly good.

Table 1. Average top-5 scores for accessory items. The database contains 650k accessory items gathered from web shopping mall.

| Original | Classifier | Ours$_1$ | Ours$_2$ |
|---|---|---|---|
| 2.44 | 2.09 | 3.26 | 3.39 |

To apply our first method (Ours$_1$) we must forward-pass all item images from shopping mall to the classifier to get the class label for one-hot augmentation. It makes difficult to be applied in industry. Alternatively, we might be able to apply text classification technic to extract class label from the item texts, which will cost much time also. Therefore, in the second experiment, we only tested the original and our second method (Ours$_2$). Word vector is extracted from product item data following the method described in section **3.4**. Test set contains 200 digital item images collected from multiple websites and customer reviews. Table 2 shows the average top-5 score. The base performance is higher than the performance in accessory database due to clear images and well-trained base network

Table 2. Average top-5 scores for digital items. The database contains 380k digital items gathered from web shopping mall.

| Original | Ours$_2$ |
|---|---|
| 2.68 | 3.48 |

## 7. Product Search Service in Industry

Nowadays, we can find similar item search in many web services and devices. One of the popular product search service is in the newest Galaxy series called Bixby. Shopping mode in Bixby supports item search in web shopping mall. It can be used through not only the camera but also the gallery. In the process, no further information is needed other than user image whereas the shopping mall database covers both text and image. Figure 1 shows an example of similar item search. The input image is a mouse device and the results are mousses with similar look. It shows that category matching is also important as visual similarity. Our methods are especially useful in these kinds of service in many industries

## 8. Discussion and Conclusion

Some of failure cases in our methods are due to the misclassification labels from the neural network classifier. Although we do not put high weight on textual vector, classification error is still a noticeable factor. Since user taken images in real world are not always well-taken or well-recognizable and even includes Moiré or severe noise under extreme environment such as low-light, it is hard to enhance the classification performance. There are several researches [30]–[33] that have proved classification in real world is not always as good as the performance in the paper, and some of them propose several methods to solve it. In future, we might apply those technics for our classifier. Thereby, we may get better results in near similar search.

Our textual vector space does not correspond to classification label space. Textual similarity score is different from the confusing score in classification network. Since our method use textual vector for class-wise search, classification label is better to be projected onto the textual space. For example, if label $k$ is highly likely to be misclassified as a label $l$, then the similarity of $k$ and $l$ is required to be high in textual space too. Unfortunately, our method did not preserve the confusing similarity from the classifier. In future work, we should further develop the method for better similarity projection

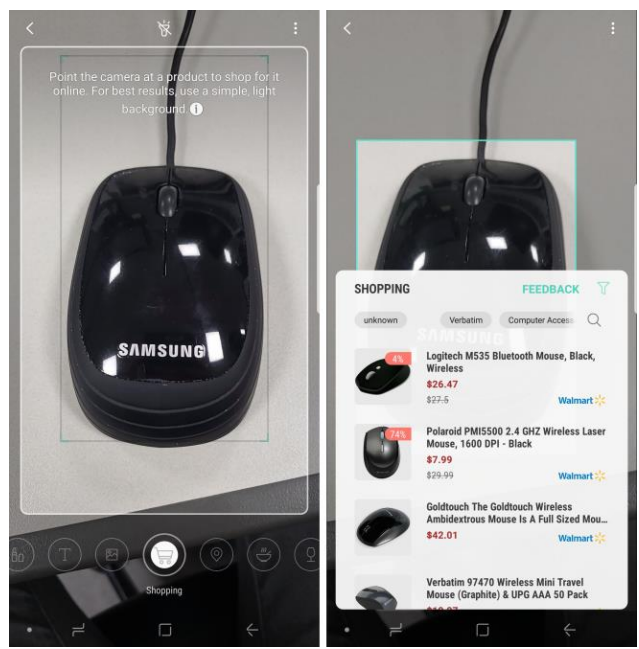

Figure 1. Bixby Shopping service in Galaxy S series. It searches similar products based on object similarity and category.